\documentclass[letterpaper, 10 pt, journal, twoside]{ieeetran}

\usepackage{cite}
\usepackage{amsmath,amssymb,amsfonts}
\usepackage{algorithmic}
\usepackage{graphicx}
\usepackage{textcomp}
\usepackage{xcolor}
\usepackage{enumerate}
\usepackage{enumitem}

\PassOptionsToPackage{hyphens}{url}\usepackage{hyperref}
\usepackage{multirow}
\usepackage{soul}
\usepackage{dirtytalk}
\usepackage{tikz}

\begin{document}

\title{Using Mobile AR for Rapid Feasibility Analysis for Deployment of Robots: A Usability Study with Non-Expert Users
}

\author{Krzysztof Zielinski$^{1,2}$, Slawomir Tadeja$^{3}$, Bruce Blumberg$^{2}$, Mikkel Baun Kjærgaard$^{1}$ %
\thanks{Manuscript received: December 13, 2024; Revised February 22, 2025; Accepted April 8, 2025.}
\thanks{This paper was recommended for publication by Editor Angelika Peer upon evaluation of the Associate Editor and Reviewers' comments.
This work was supported by Innovation Fund Denmark.} 
\thanks{$^{1}$K. Zielinski and M. Kjærgaard are with Faculty of Engineering, University of Southern Denmark, Denmark
        {\tt\footnotesize krzi, mbkj@mmmi.sdu.dk}}%
\thanks{$^{2}$K. Zielinski and B. Blumberg are with Universal Robots A/S, Denmark
        {\tt\footnotesize krzi, brbl@universal-robots.com}}%
\thanks{$^{3}$S. Tadeja is with Department of Engineering, University of Cambridge, UK
        {\tt\footnotesize skt40@cam.ac.uk}}%
\thanks{Digital Object Identifier (DOI): see top of this page.}
}


\markboth{IEEE Robotics and Automation Letters. Preprint Version. Accepted April 2025}
{Zielinski \MakeLowercase{\textit{et al.}}: Using Mobile AR for Rapid Feasibility Analysis for Deployment of Robots} 

\maketitle

\begin{abstract}

Automating a production line with robotic arms is a complex, demanding task that requires not only substantial resources but also a deep understanding of the automated processes and available technologies and tools. Expert integrators must consider factors such as placement, payload, and robot reach requirements to determine the feasibility of automation. Ideally, such considerations are based on a detailed digital simulation developed before any hardware is deployed. However, this process is often time-consuming and challenging. To simplify these processes, we introduce a much simpler method for the feasibility analysis of robotic arms' reachability, designed for non-experts. We implement this method through a mobile, sensing-based prototype tool. The two-step experimental evaluation included the expert user study results, which helped us identify the difficulty levels of various deployment scenarios and refine the initial prototype. The results of the subsequent quantitative study with 22 non-expert participants utilizing both scenarios indicate that users could complete both simple and complex feasibility analyses in under ten minutes, exhibiting similar cognitive loads and high engagement. Overall, the results suggest that the tool was well-received and rated as highly usable, thereby showing a new path for changing the ease of feasibility analysis for automation.
\end{abstract}


\begin{figure}[h]
    \begin{center}
        \includegraphics[width=.6\columnwidth]{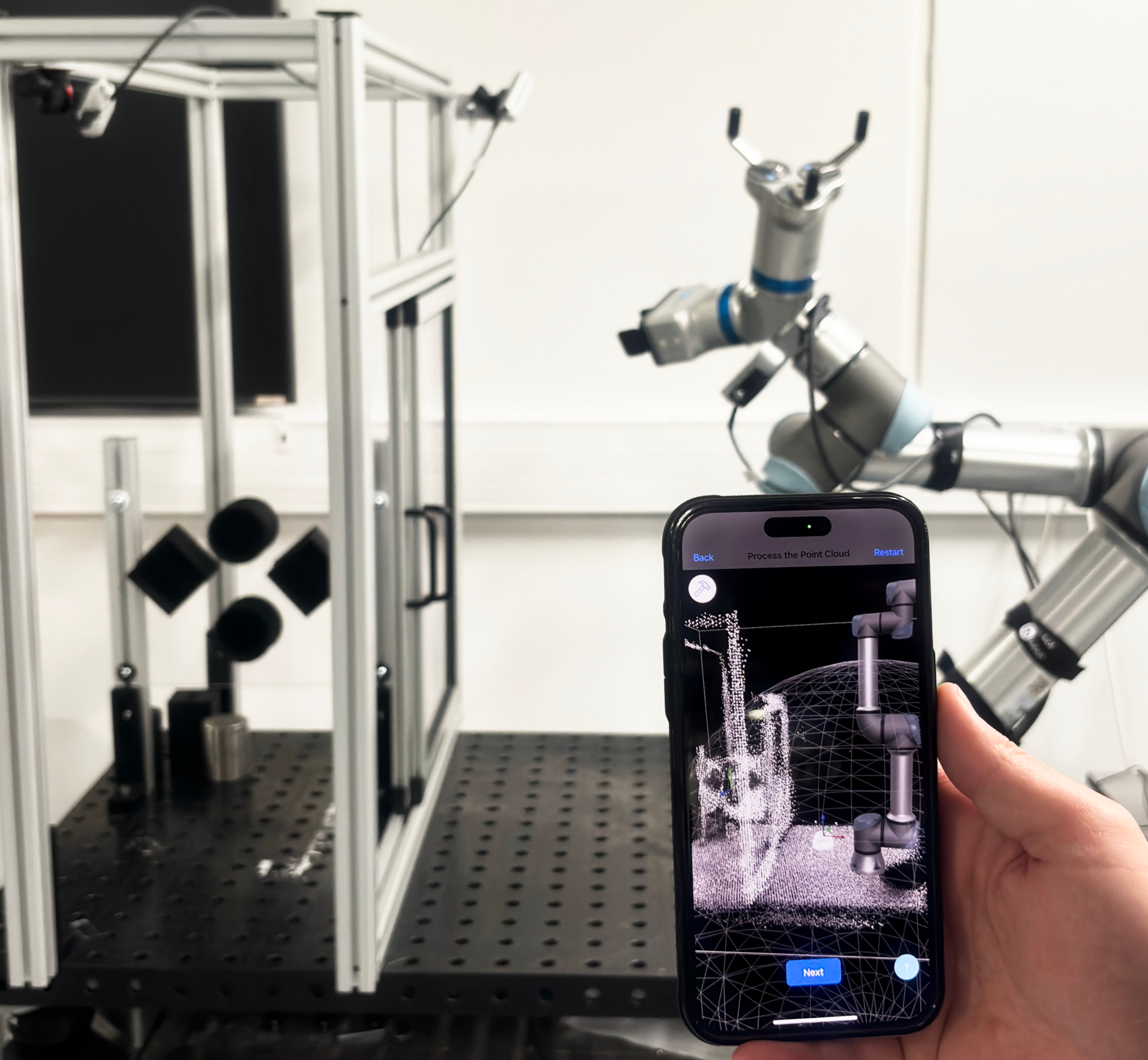}
        \caption{Evaluating the feasibility of the machine loading automation.}
        \label{fig:machine_loading}
    \end{center}
\end{figure}

\section{Introduction}
\IEEEPARstart{I}{n} recent years, manufacturers have increasingly considered automating production lines with collaborative robots (cobots) to enhance efficiency, accuracy, and cost-effectiveness~\cite{keshvarparast_collaborative_2024}. This shift is driven by labor shortages, particularly affecting Small and Medium Enterprises \cite{ajewole_unlocking_2023}.

The initial step in automation is conducting a feasibility analysis to determine the viability of robotic automation~\cite{hetmanczyk_method_2024}. This involves addressing critical questions such as: \textit{Which robot and gripper should be selected? What are the safety considerations? What cycle time is required?} Answering these often requires extended downtime of current production lines, leading to potential revenue losses.

Converting a production line to an automated one poses a significant financial risk without adequate feasibility studies. Integrators specializing in the deployment of automated systems often use their expertise to estimate the feasibility of automation~\cite{website:robotics247}. While quick and cost-effective, this approach can lack data validation and may miss important aspects of the final configuration.

Offline simulation using digital or virtual twins offers a more thorough approach that avoids long downtimes~\cite{thelen_comprehensive_2022}. However, creating such models incurs high financial and temporal costs, as designers need to capture all physical characteristics of the space, recreate them virtually, and develop complex simulations~\cite{bing_digital_2024}. {\color{blue}Industrial tools such as RoboDK~\cite{website:roboDK} and Visual Components~\cite{website:visualComponents} target expert users in 3D simulation and robotics.} While this might justify the high initial costs for an operational workcell, it may not be worth it for a feasibility analysis.

To address these challenges, we propose using a mobile \textit{augmented reality} (AR) interface \cite{tadeja_mobileAR_2023} for a rapid feasibility analysis (see \autoref{fig:machine_loading}). The AR-based tool enables non-expert users to quickly capture and modify a 3D model of the current workcell directly on a mobile device~\cite{zielinski_precise_2024}. It uses a virtual robotic arm representation to help users understand how well it fits in their workcell. Users can also input semantic data about their workcell~\cite{zielinski_robotgraffiti_2024}, which is then used in a path planning pipeline to determine if the setup is a feasible automation solution. The entire process is designed to be simple and fast, allowing users to obtain results in under ten minutes, as shown in our user studies. We conducted two user studies: an expert user study to understand users' pain points and differences in difficulty for various automation tasks, and a quantitative study to evaluate usability with a larger sample of non-expert users.
In summary, we claim the following contributions: 1) Rapid, mobile AR feasibility analysis tool for reachability study of a robotic arm requiring minimal training; 2) Expert user study to identify difficulty levels in automation tasks.; 3) Results of a quantitative user study with 22 non-expert users, assessing cognitive load, engagement, and success rates in two automation tasks.

\section{Related work}
The deployment of robotic systems involves comprehensive feasibility analysis, which encompasses both technical and economic aspects~\cite{weidemann_literature_2023}. Moreover, understanding concepts such as reachability is crucial for optimizing performance and ensuring cost-effectiveness. Here, we review the existing literature on reachability analysis and feasibility studies, highlighting key research that informed our approach. {\color{blue}Additionally, we explore methodologies for human-robot interaction (HRI) user studies from \cite{leichtmann_crisis_2022}.}

\subsection{Reachability Analysis in Robotics}
Feasibility analysis in robotics often involves reachability analysis, which is crucial for defining waypoint requirements and determining the optimal base placement of robots~\cite{kostal_reachability_2017}. The \textit{Reachability Maps} (RMs) and their derivatives have been extensively studied. Traditionally, RMs are generated using \textit{Inverse Kinematics} (IK) solvers to identify possible joint-space configurations~\cite{vahrenkamp_representing_2015}. However, IK solvers are not the most efficient method for generating RMs, leading to the development of alternative approaches leveraging machine learning. One such approach is \textit{ReachNet}, a neural network-based algorithm designed for rapid RM generation, enabling real-time applications~\cite{sandakalum_reachnet_2023}. Whereas, Gienger et al.~\cite{gienger_data-based_2024} utilized a \textit{Random Forest} algorithm to accelerate RM generation, demonstrating its effectiveness in large-scale construction planning. Additionally, Yao et al.~\cite{yao_enhanced_2024} introduced the concept of \textit{Extended Dexterity Maps} by constraining RMs through the evaluation of disjoint flip solutions.

While RMs address feasible reach locations, \textit{Inverse Reachability Maps} (IRMs) are used to determine the optimal base positions for robots to perform specific tasks. An example of this approach is \textit{Reuleaux}, which employs IRMs for defining feasible robot base locations~\cite{makhal_reuleaux_2018}. Furthermore, Weingartshofer et al.~\cite{weingartshofer_optimal_2021} developed a joint-space path planner to create IRMs, optimizing both the robot base and Tool Center Point positions. The generation of multiple RMs to enhance robot placement flexibility was explored in industrial settings, as described in a relevant patent~\cite{prats_generation_2020}.

Reachability analysis is also vital for \textit{Mobile Manipulation} (MM) systems, where it helps define the placement of mobile platforms for specific tasks. Jauhri et al.~\cite{jauhri_robot_2022} generated offline IRMs to query potential solutions for optimal mobile platform placement. Similarly, Sustarevas et al.~\cite{sustarevas_task-consistent_2021} created task-specific IRMs for mobile 3D printing applications. IRMs are also applicable in determining optimal grasp positions~\cite{wang_optimal_2021}.

\subsection{Feasibility Analysis in Robotics and Automation}
Feasibility analysis is a crucial component in the deployment of robotic systems, typically divided into technical and economic feasibility~\cite{weidemann_literature_2023}. Studies have shown that economic feasibility often has a more significant impact~\cite{kozul-wright_industrial_2017}, particularly since automating production lines, such as those involving \textit{Computer Numerical Control} (CNC) machines, can be prohibitively expensive~\cite{soori_robotical_2024}. To address this, various models have been developed to evaluate the economic feasibility of robotic automation~\cite{vido_collaborative_2024}. These analyses are also instrumental in assessing the maturity of robotic systems within organizations~\cite{hetmanczyk_method_2024}.
At the same time, efforts to enhance technical feasibility primarily focus on simplifying the programming of robotic arms. Giberti et al.~\cite{giberti_methodology_2022} propose a skill-based programming approach to streamline deployment. Similarly, Pedersen et al.~\cite{pedersen_robot_2016} advocate for task-level programming, where users can utilize task-specific, pre-defined sets of instructions to create more complex programs.

\begin{figure*}[th]
    \begin{center}
        \includegraphics[width=\textwidth]{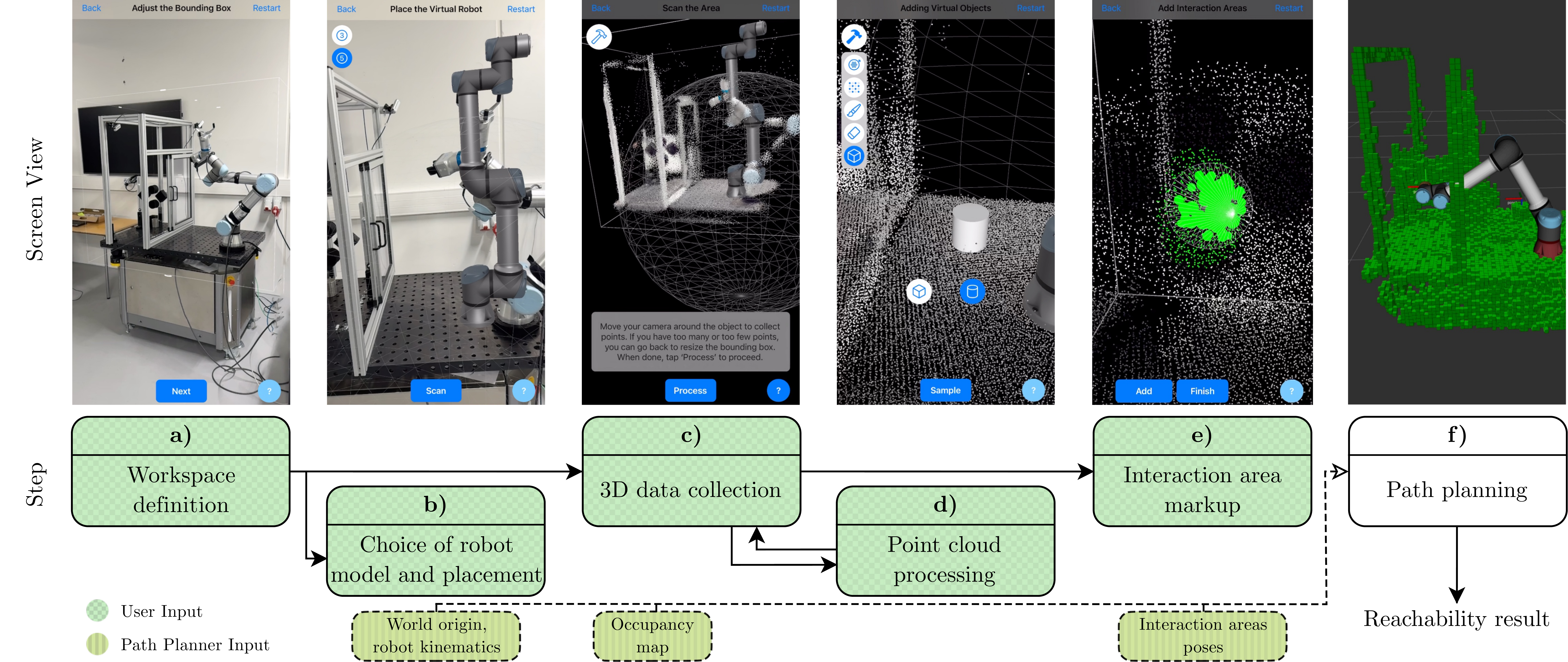}
        \caption{Proposed method workflow: a) workspace definition using bounding box; b) choice of robot model and placement; c) point cloud generation; d) point cloud processing: outlier removal, downsampling, brush eraser, sponge eraser, object addition; e) interaction area markup; f) path planning. {\color{black}Steps a-e) require user input and are completed on the user's device. Step f) is triggered by providing inputs from steps b), c) and e) and generates reachability results in the cloud. \textit{Best practices to obtain good scan is ``free play''}: The user can always return to the previous step, make the bounding box larger or smaller as needed, move around to get a different perspective and use the tools to enhance the point cloud.}}
        \label{fig:workflow}
    \end{center}
\end{figure*}

\subsection{Role of User Studies in Human-Robot Interaction}
HRI relies on user studies to evaluate the effectiveness and usability of robotic systems. The need for reliable and replicable results in these studies is paramount~\cite{leichtmann_crisis_2022}. Rajendran et al.~\cite{rajendran_framework_2020} offer a framework for conducting HRI studies that emphasize replicability and reliability. Fraune et al.~\cite{fraune_lessons_2022} provide lessons learned from HRI experts on designing and conducting user studies, stressing the importance of methodological rigor.

User studies are also a common and desirable evaluation approach in AR \cite{10.3389/frobt.2018.00037}. They are essential for determining the usability of AR applications, which is crucial for their adoption success~\cite{xiang_user_2022}. However, many AR user studies are conducted under inadequate experimental conditions and, therefore, lead to misleading results~\cite{zhang_should_2024, wozniak_enhancing_2024}. Thus, the correct experimental design is crucial for generating high-quality results. Researchers utilize user studies to evaluate AR interactions with robotic systems, such as assessing collection interfaces for robot learning~\cite{jiang_comprehensive_2024} or configuring safety zones~\cite{cogurcu_augmented_2023}. For instance, Tadeja et al.~\cite{tadeja_enhancing_human_robot_collaboration_with_arpdf_2024} conducted user studies with an AR-based system to enhance human-robot collaboration through both formative and quantitative evaluations.


In our approach, we aim to empower expert and non-expert users alike at the pre-deployment phase by helping them make educated decisions about introducing robots into their production lines (see \autoref{fig:machine_loading}). We achieve this by enabling rapid feasibility analysis with minimal costs based on reachability studies employing a context-aware path planner. {\color{blue}Recognizing the ongoing need for research in AR-supported HRI, this paper evaluates an AR interface for robot reachability assessment using established HRI methodologies, providing insights into the application of how AR can support the crucial initial phase of robot integration within human environments.
}

\section{Rapid Feasibility Analysis}
We introduce a novel tool for rapid feasibility analysis of a robotic arm's reachability (see \autoref{fig:machine_loading}). We have built the tool based on our previous iterations~\cite{zielinski_precise_2024, zielinski_robotgraffiti_2024} that focused on a technical solution by utilizing AR for optimal robot deployment. {\color{black}We have improved and extended our previous system~\cite{zielinski_robotgraffiti_2024} by introducing point cloud manipulation tools~\cite{zielinski_precise_2024} and path planning capability enabled by environment awareness. Moreover, we have} previously primarily focused on technical implementation but did not consider the end-to-end solution at the usability level. Therefore, we address this gap by conducting comprehensive user studies to evaluate the usability and overall effectiveness of the tool in real-world scenarios. User studies are crucial for defining the usability of the system and ensuring a human-centered focus for meaningful interaction~\cite{apraiz_evaluation_2023}.

The developed tool supports reachability analysis by breaking it down into six steps (see \autoref{fig:workflow} {\color{black}and video}\footnote{\url{https://youtu.be/lCmHNPy0L-o}}):

\textbf{Defining Workspace}: The user sets the extent of a virtual bounding box to define the environment. The workspace should encompass the automated area, such as for the machine loading task: the opening of the machine, the fixture for the workpiece (like a chuck), and the area where the robot should be positioned. The user can opt to include collision areas, like a wall, in the extended area.
\textbf{Choosing Robot Model and Placement}: {\color{blue}The user selects the robot size and \say{places} a virtual robot on a physical or virtual flat surface.}
An initial reachability sphere associated with a given robot model is displayed to guide users regarding size and placement requirements. Note that this reachability sphere is, at best, a gross approximation of the actual configuration space and does not account for obstacles or other constraints (e.g., singularities), {\color{black}and is only intended as a visual help for the user.}
\textbf{3D Data Collection}: The user collects a point cloud representation of the environment within the bounding box.
\textbf{Point Cloud Processing Toolbox}: {\color{black}We assume that our solution will be used by users who do not know best practices for 3D scanning, and therefore, our workflow is designed to be iterative and offers a set of tools allowing} to modify the captured scan: (i) \textit{outlier removal} i.e., noise removal from the point cloud; (ii) \textit{downsampling}, i.e. reduction of points in the point cloud; (iii) \textit{brush eraser}, i.e., point eraser using a cone-shaped brush on the screen; (iv) \textit{sponge eraser}, i.e., point eraser using the device as a sponge to \say{wipe off} points, and (v) \textit{object addition}, i.e., adding simple primitives to the scan.
\textbf{Interaction Area Markup}: The user marks points with a cone-shaped green spray {\color{black}and defines the robot's approach direction using the device's orientation relative to the marked area.}The user defines as many interaction areas as necessary, such as workpiece source and sink areas and the machine workpiece fixture area, for a machine loading task.
\textbf{Path Planner}: The collected data is used to run an online path planner that attempts to reach the specified interaction areas with the robot. The planner employs point cloud data converted into a 3D occupancy map for collision detection and defined robot kinematics to avoid singularities.

The result of the last step is a successful or unsuccessful reachability study based on the user input. In case of failure, the user can modify the captured data without the need to restart the entire process.

\subsection{Tool Implementation}
We implemented our prototype of the tool using iPhone 15 Pro Max running iOS 17.6.1 for steps a-e, as shown in \autoref{fig:workflow}. {\color{black}For the path planning step, we utilized MoveIt2 with the single-query LBKPIECE path planner}~\cite{sucan_kinodynamic_2010} in ROS2 Indigo. {\color{black}The chosen planner prioritized efficiency and robustness over the optimality of the solution.}

The data collected from the mobile device (point cloud, robot size and placement, interaction areas) are published as ROS2 topics. Moreover, the generated point cloud is converted into an Octomap representation, the robot placement is used as the origin of the simulation, and the interaction zones are set as target poses for planning.

\section{Expert User Study}
As the first stage of our user study, we conducted a user study with domain experts. This stage focused on understanding user needs in-depth and gathering insights into automation task complexities. All these experts were \textit{application engineers} (AEs) who specialize in defining automation feasibility with experience ranging from a minimum of 5 up to 35 years. Based on this study, we were able to ascertain the following aspects of our approach:
    (i) time required to complete the feasibility analysis; and (ii) difficulty levels of different automation tasks.



\subsection{Tasks}
Participants performed a feasibility analysis of a cobot for two automation tasks using both the proposed tool and RoboDK, a widely-used industrial automation simulator. The tasks were:

\begin{enumerate}[label=(\textbf{\Alph*}), wide=0pt]
    \item pick \& place on a flat surface: Participants were provided with a flat workspace, a table, and two designated areas on the table: the workpiece pick area and the place area. They were then asked to select a suitable robot placement within the workspace, considering the available virtual robot model.
    \item machine loading (a process of placing and removing workpieces into or out of a machine, e.g., CNC for machining): Participants were given a workspace that included a table and a mockup of a CNC machine (as seen in \autoref{fig:machine_loading}). They were also provided with two marked areas: the workpiece source on the table and the workpiece fixture in the CNC machine on the vertical wall. The task required them to choose a robot placement on the table based on the available virtual robot model.
\end{enumerate}



{\color{blue} We have opted not to include end-effectors in our tasks, as they do not increase user complexity and our focus is on conducting a usability study rather than developing an industry-ready product. However, we plan to add them in the future to account for reachability differences due to the tool flange offset. Adding a gripper will likely limit the reachability region around pick-and-place areas by reducing possible gripping orientations. It will also likely affect the reachability region during arm movement, as the end-effector offset will necessitate different trajectories to avoid collisions.}

\subsection{Study Protocol}
\label{sec:formative_study:protocol}
The study began with a background interview to verify participants' expertise. Next, the experts were assigned the \textbf{(A)} and \textbf{(B)} tasks. They started by assessing the difficulty of each task using their usual tools and methods on a 5-point Likert-like scale. Subsequently, experts performed the simpler task (as per their initial assessment) using the offline simulation software, RoboDK{\color{black}, aiming to reduce the test duration.}

Following this, the experts were introduced to our mobile AR prototype via a brief video demonstrating its functionalities. They then completed both tasks using the proposed method in a randomized order. 
The study concluded with a semi-structured interview to collect expert insights on the prototype and suggestions for further improvement. To complete the data gathering, we collected the time it took the participants to complete both tasks using the proposed method, and the easier task using RoboDK (see \autoref{tab:formative_time}).

\begin{table}[htbp]
\caption{Average completion time of feasibility analysis of tasks \textbf{(A)} \& \textbf{(B)} using proposed method and RoboDK.}
\label{tab:formative_time}
\begin{center}
\begin{tabular}{c|c|c}
\textbf{Task} & \textbf{Method}                  & \textbf{Average completion time} \\
              &                                  & {\color{blue}{[}min:sec{]}}                        \\ \hline
A             & \multirow{2}{*}{Proposed method} & {\color{blue}7:14}                             \\
B             &                                  & {\color{blue}6:56}                             \\ \hline
A             & RoboDK                           & {\color{blue}25:49}                            
\end{tabular}
\end{center}
\end{table}

\section{Expert User Study Results}


\subsection{Task Complexity}

Experts consistently indicated that task \textbf{(A)}, i.e., pick \& place on a flat surface, was easier than task \textbf{(B)}, i.e., machine loading. This was reported both before and after using our prototype, as shown in \autoref{fig:formative_difficulty}. However, the difficulty of both tasks decreased after using the prototype, by 4\% and 20\% respectively. Furthermore, the gap in difficulty levels between tasks \textbf{(A)} and \textbf{(B)} narrowed from 28\% to 14\%, suggesting that the proposed method is a more effective tool for both simple and complex tasks compared to existing methods.

\subsection{Task Completion Times}
\begin{figure}[htbp]
    \begin{center}
        \includegraphics[width=.99\columnwidth]{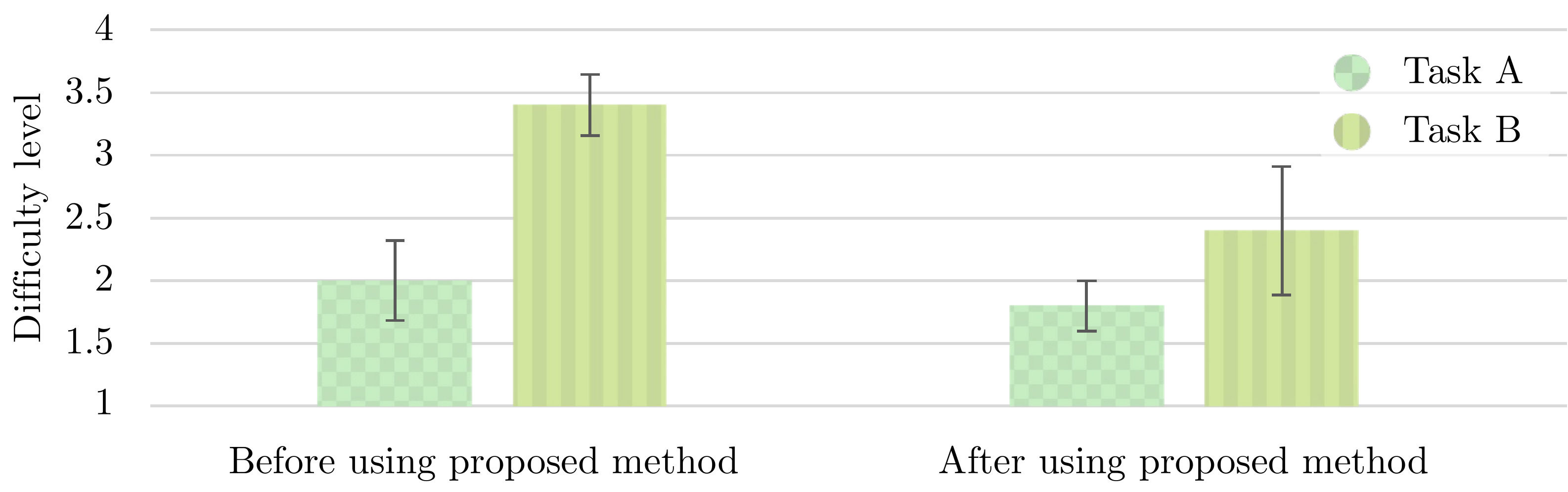}
        \caption{Chart of the difficulty level of \textbf{(A)} and \textbf{(B)} tasks on 5-point Likert-like scale, as assessed before and after using our prototype with whiskers denoting standard deviation error.}
        \label{fig:formative_difficulty}
    \end{center}
\end{figure}

As seen in \autoref{tab:formative_time}, both tasks were completed on average in {\color{blue}7:05} [min], a significant improvement over RoboDK. Additionally, we set a 30 [min] cutoff time for participants to complete the task using RoboDK, and only two participants managed to do so. To further highlight the difference, the experts who completed the task using RoboDK only managed to model the workcell, skipping the path planning part. All participants attempted to complete task \textbf{(A)} using RoboDK, as it was perceived as the easier task, as indicated in \autoref{fig:formative_difficulty}.



\vspace{-5mm}

\subsection{Prototype Refinement}
Based on the results of the expert user study, participant observations, and semi-structured interviews, {\color{black}we observed that participants initially struggled with setting a bounding box and understanding the descriptions of the available tools. Consequently,} we have applied the following refinements:
\begin{enumerate}
    \item \textbf{Clear Instructions:} We enhanced user guidance through the available steps and simplified the step-by-step process, creating a more linear flow. Additionally, each tool in the processing toolbox now features more straightforward phrasing with clear calls to action.
    \item \textbf{Improved Bounding Box Manipulation:} Resizing handles have been reworked to scale based on the user's distance to the virtual box, making them easier to press.
    \item \textbf{Improved Pipeline:} {\color{blue}We improved ROS2 node implementation for smooth data handover to MoveIt2.}
\end{enumerate}

Furthermore, we utilized experts' evaluations of the difficulty levels of automation tasks (see \autoref{fig:formative_difficulty}) to design our quantitative user study with non-expert users. In this subsequent study, we aimed to determine if our method can reduce the difficulty for non-expert users.

\section{Quantitative User Study}
In addition to implementing the changes informed by the expert user study, we designed the second stage of the user study with a larger participant sample size to answer the following questions: \textit{Can the method be used by non-experts? Can non-expert users carry out feasibility analysis with little to no deployment experience? How quickly can they complete the tasks? Can the method be used for tasks with varying complexity levels?}

\subsection{Participants}
For this iteration of the study, we used opportunity sampling to recruit 22 participants. They ranged in age from 21 to 58 years old and included 12 males, 9 females, and a single participant who chose not to disclose their gender. Twenty participants had some prior experience with AR applications, such as IKEA or Pokémon GO. The participant pool included a mix of undergraduate and graduate students, professors, and working professionals.

\subsection{Study Protocol}
We began the experiment by collecting relevant background information to ensure that participants had minimal experience in robot deployment. Participants were then introduced to the concept of robot automation and the necessity of feasibility studies. Subsequently, they were presented with two tasks of varying difficulty levels, as identified by experts in the expert user study (see \autoref{fig:formative_difficulty}): \textbf{(A)} pick \& place on a flat surface (simple) and \textbf{(B)} machine loading (complex).

Next, we provided a brief overview of our prototype and its functionalities (see \autoref{fig:workflow}). Participants then proceeded to carry out the first task. The order of task execution was fully balanced to mitigate the effects of transfer learning. 
After completing each task, participants filled out several questionnaires, i.e., the \textit{NASA Task Load Index} (TLX)~\cite{hart_1988_development} to measure their perceived cognitive load when carrying out each task, the \textit{Flow Short Scale} (FSS)~\cite{rheinberg2003} to assess flow and anxiety, and the \textit{System Usability Scale} (SUS)~\cite{brooke2013} to ascertain perceived usability of the prototype.

\begin{figure}[htbp]
    \begin{center}
        \includegraphics[width=.99\columnwidth]{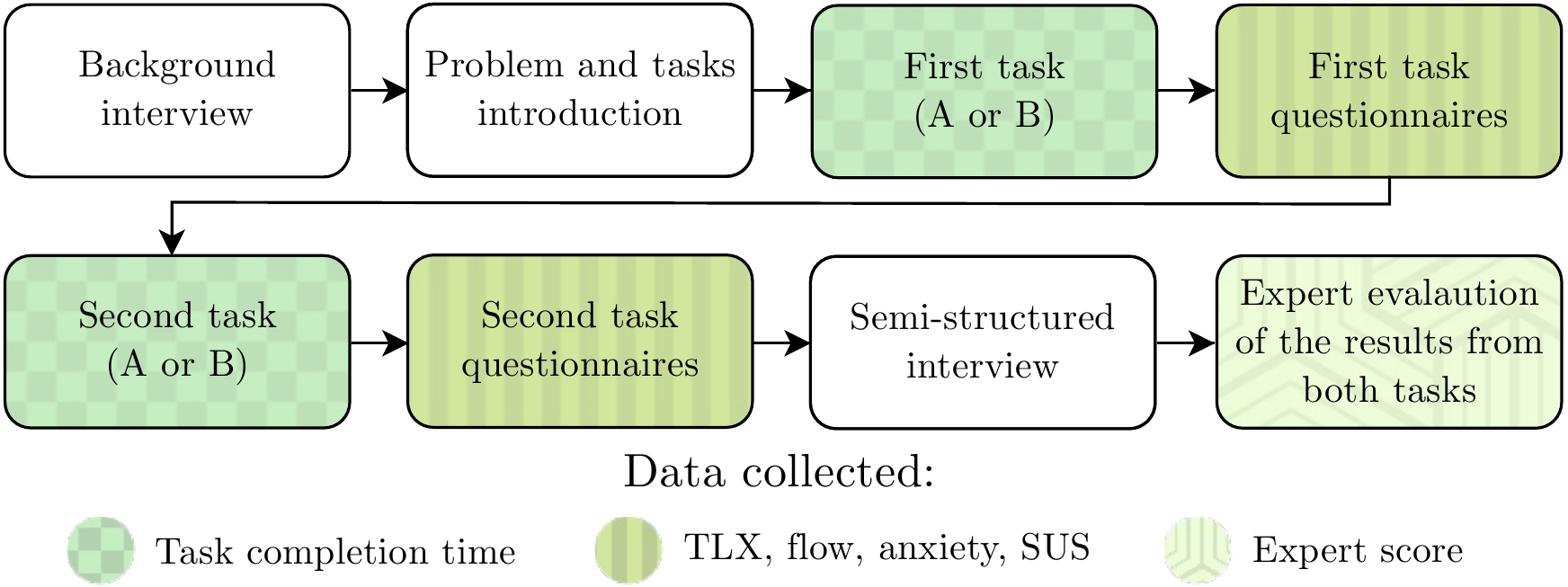}
        \caption{Flow chart of quantitative user study design.}
        \label{fig:flow_quantitative}
    \end{center}
\end{figure}

Finally, participants took part in a semi-structured interview to discuss their likes and dislikes with respect to the proposed approach. We also collected the task completion times to compare them against the experts' results. The flow of the user study can be seen in \autoref{fig:flow_quantitative}.

Additionally, we asked three experts from the expert user study to evaluate the obtained trajectory for each task from all participants. The results can be seen in \autoref{tab:quantitative_fss_sus}. The experts assigned scores based on the following criteria:
\begin{enumerate}[label=\textbf{\arabic*} --,wide=0pt]
    \setcounter{enumi}{-1}
    \item The collected scan is not useful for feasibility analysis (e.g., insufficient detail/too much noise for collision avoidance, interaction zones are incorrectly marked).
    \item The collected scan might allow a feasibility study (e.g., it is possible to identify if the robot can reach interaction zones, but path planning cannot be checked due to poor quality).
    \item The collected scan can be used for feasibility analysis (e.g., the scan is accurate enough for collision avoidance, interaction zones are properly marked, and the path planner can solve the solution).
\end{enumerate}

\section{Quantitative User Study Results}
\subsection{Task Completion Times}
The average completion times for tasks \textbf{(A)} and \textbf{(B)} were {\color{blue}6:53} [min] and {\color{blue}7:40} [min], respectively, as seen in \autoref{tab:quantitative_fss_sus}. For statistical analysis, we used a natural log transformation to normalize the distribution. The Shapiro-Wilk test for normality failed to reject the null hypothesis ($W = .972, p = .359$), allowing us to use paired t-tests to check for differences between the two tasks. The test revealed no statistical significance between the tasks ($t(21) = 1.376, p = .183$). Therefore, we have also used paired Two One-Side t-tests (t-TOST) to test for equivalence. Based on the t-TOST test ($p = .0002$), we rejected the presence of effects more extreme than {\color{blue}-30 to 30 [sec]} (a range chosen based on the desired goal of task completion: 10\% of 10 minutes), suggesting that the level of difficulty of the more complex task \textbf{(B)} did not significantly impact the time required to complete it using the proposed method within equivalence interval. Moreover, these times closely align with the experts' average result of {\color{blue}7:05} [min] (\autoref{tab:formative_time}).

\subsection{Expert Result Evaluation}
We asked three experts to evaluate the participants' feasibility results based on their robot deployment experience. For task \textbf{(A)}, the average score was 1.68, with no participant scoring below 1. In contrast, for the task \textbf{(B)}, the average score was 1.34, with only four participants scoring below 1.

Typical errors identified by the experts included incorrectly marked poses (interaction areas) and missing critical elements of the workcell. The lower scores for task \textbf{(B)} were primarily due to a higher number of errors related to missing elements, such as unscanned tops of machines or walls. These issues were often caused by improperly set up bounding boxes that failed to capture all necessary elements.

However, it is important to note that all participants, except one, generated scans that could be used to plan at least one of the marked interaction areas. This indicates that despite some errors, the majority of participants were able to produce results that were sufficient to carry out feasibility studies.

\subsection{Questionnaires}
\begin{figure}[htbp]
    \begin{center}
        \includegraphics[width=.99\columnwidth]{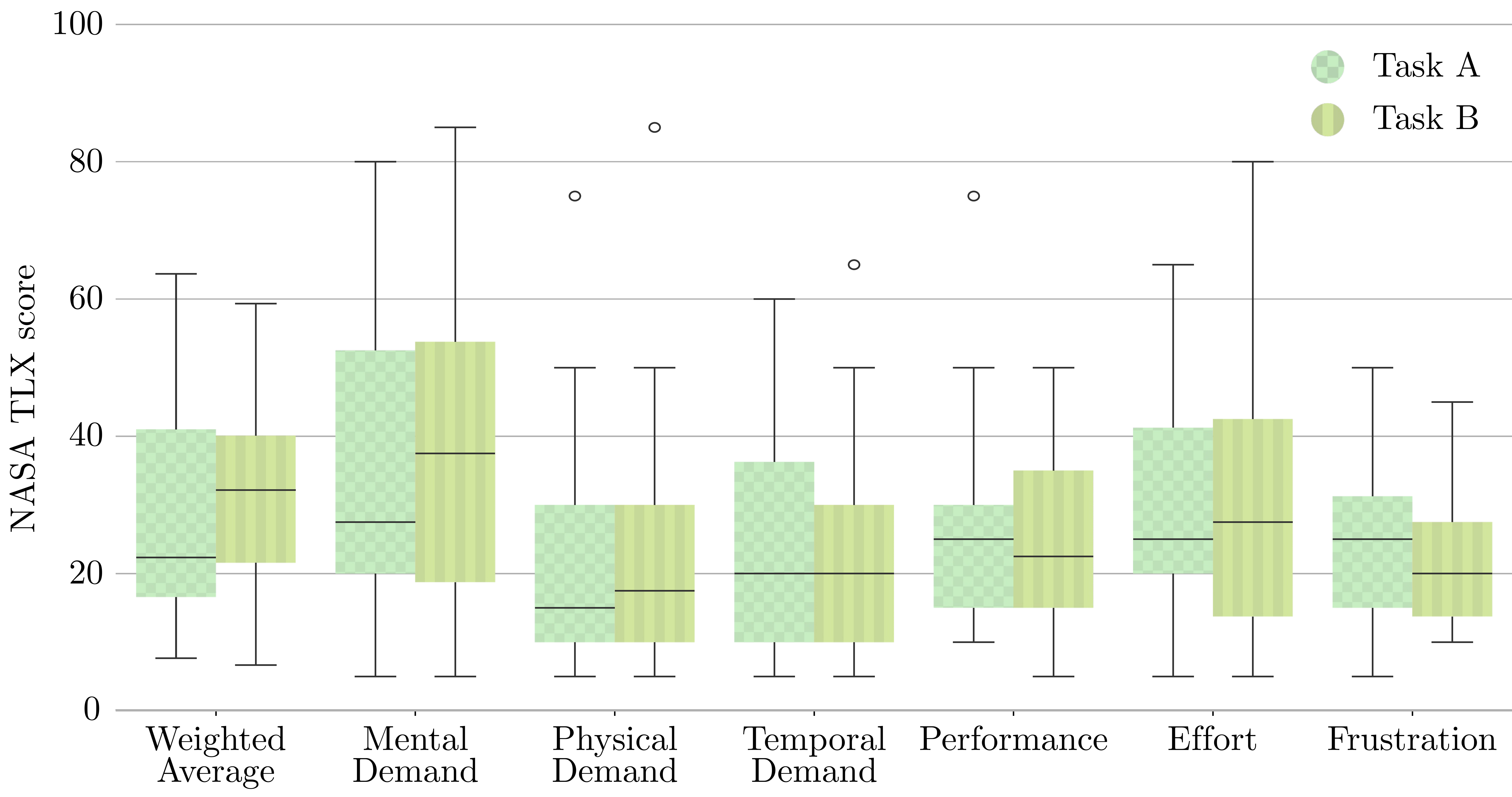}
        \caption{Cumulative results of TLX survey of tasks \textbf{(A)} and \textbf{(B)}. The weighted average is calculated using pairwise comparisons. Lower scores are better (0--low demand/good performance, 100--high demand/bad performance).}
        \label{fig:quantitative_nasa}
    \end{center}
\end{figure}

The cumulative scores in the TLX questionnaires are shown in \autoref{fig:quantitative_nasa}. The weighted average for task \textbf{(A)} was 30.2, and 31.72 for task \textbf{(B)}, which can be both considered \say{somewhat high}, slightly crossing the bottom threshold of 29 (\autoref{tab:quantitative_fss_sus})~\cite{Prabaswari_2019_NASA_TLX_Table}. 


The Shapiro-Wilk test showed deviation from normality ($W = .942, p = .04$), so we used the Wilcoxon signed-rank test to compare tasks \textbf{(A)} and \textbf{(B)}. The results ($W = 83.0, p = .629$) indicated no significant difference in TLX results between the tasks. We also employed Wilcoxon TOST with equivalence region of -10 to 10, that rejected null hypotheses ($p = 0.013$), indicating equivalence within the interval. This shows that task complexity has little to no effect on cognitive load when using the tool for feasibility studies.

Next, we analyzed the FSS results, which showed an average flow level of 4.96 for task \textbf{(A)} and a slightly higher value of 5.19 for the more complex task \textbf{(B)}, and an average anxiety level of 3.43 for task \textbf{(A)} and 3.82 for task \textbf{(B)}, as seen in \autoref{tab:quantitative_fss_sus}. For flow level, the Shapiro-Wilk test for normality did not reject the null hypothesis ($W = .969, p = .269$), allowing us to use a paired t-test to test for differences between tasks \textbf{(A)} and \textbf{(B)}. The test revealed no statistical significance between the two tasks ($t(21) = 1.346, p = .193$). Hence we used t-TOST to test for equivalence of the results. For the range of -0.35 to 0.35 (chosen based on 10\% of the FSS scale), t-TOST failed to reject one of the hypotheses ($t1(21) = .686, p = .25, t2(21) = -3.378, p = .001$), therefore the results are inconclusive.
For anxiety level, the Shapiro-Wilk test for normality did not reject the null hypothesis ($W = .959, p = .118$) and paired t-test found significant differences ($t(21) = -2.306, p = .031$) between the samples. This suggests that the task complexity level affects the task flow and anxiety levels.

Finally, the SUS scores (\autoref{tab:quantitative_fss_sus}), which returned average scores of 74.37 for the task \textbf{(A)} and 75.92 for the task \textbf{(B)}, both above the interface usability threshold of 70. At the same time, these results suggest that our tool needs to be further refined to offer an optimal interface for non-experts.

\begin{table}[htbp]
\caption{The results of the questionnaires, the completion time and expert evaluation score for quantitative user study.}
\label{tab:quantitative_fss_sus}
\begin{center}
\begin{tabular}{lc|cccc}
                       &             & \multicolumn{2}{c}{\textbf{(A)}} & \multicolumn{2}{c}{\textbf{(B)}} \\
                       &             & M              & SD            & M              & SD            \\ \hline
\textbf{TLX}           & {[}0-100{] $\downarrow$} & 30.2           & 16.57         & 31.72          & 14.5          \\
\textbf{Flow level}    & {[}0-7{] $\uparrow$}   & 4.96           & 0.68          & 5.19           & 0.76          \\
\textbf{Anxiety level} & {[}0-7{] $\downarrow$}   & 3.43           & 1.16          & 3.82           & 1.25          \\
\textbf{SUS score}     & {[}0-100{] $\uparrow$} & 74.37          & 12.24         & 75.92          & 13.37         \\
\textbf{Time}          & {\color{blue}{[}min:sec{]}}   & {\color{blue}6:53}           & {\color{blue}2:32}          & {\color{blue}7:40}           & {\color{blue}2:43}         \\
\textbf{Expert score}  & {[}0-2{] $\uparrow$}                        & 1.68           & 0.61          & 1.34           & 0.69  
\end{tabular}
\end{center}
\end{table}

\subsection{Observed Behavior and Feedback}
Multiple participants faced initial challenges with the bounding box setup, often requiring multiple attempts to set it up correctly. Issues with the bounding box handles were a frequent obstacle, with several participants struggling to manipulate them effectively. These could also lead to slightly decreased but still high average usability scores. Despite these challenges, participants enjoyed the tool’s functionalities, such as 3D scanning and point selection using spraying, and praised its visual clarity and responsiveness.

However, several areas for improvement were identified. Participants suggested that the process of adding points for pick and place could be more intuitive, and the orientation arrows for defining the robot's approach could be simplified. The need for better guidance and more informative feedback was a recurring topic, with suggestions for clearer instructions and more intuitive icons. Additionally, some participants noted that the flow of the tool could be improved to prevent users from getting lost. Enhancements such as automatic downsampling, better feedback mechanisms, and the ability to modify orientations after the initial setup were also recommended to improve the overall user experience.

Moreover, many users did not utilize all the available tools in the point cloud processing step. While some were interested in exploring all the tools, most did not see the need to erase points or downsample. This behavior can be tentatively explained by a short phase of familiarisation with the used tool or inexperience in feasibility studies that can prevent the user from exploring all available functionalities.

\section{Discussion}
The expert and quantitative user studies show strong positive results, indicating significant improvement with the proposed method over conventional approaches.

Firstly, the proposed method achieved a substantial time reduction compared to RoboDK. When using the tool, the average completion time was {\color{blue}7:16} [min] for feasibility analysis by non-expert users, whereas RoboDK usage required at least {\color{blue}25:49} [min] by an expert user.

Secondly, the quantitative user study revealed no significant difference in cognitive load across tasks of varying complexity levels, as defined by expert users. {\color{black}Therefore, our method may help to maintain a consistent cognitive demand among tasks of varying complexities.} While the results for flow levels were inconclusive, the anxiety levels indicated that users found task \textbf{(B)} more anxiety-inducing. This is likely due to the perceived increase in task complexity.


Furthermore, based on expert evaluations, non-expert users generated mostly valid feasibility analyses, meaning their results could be used as evidence of whether automation is feasible. We also observed strong transfer learning, where some participants completed the feasibility analysis in half the time of their first task, regardless of the tasks order. This suggests that users can quickly learn the tool and embrace its functionalities.

{\color{black}At the same time, the proposed method does not yield an optimal feasibility solution. Instead, our approach offers a well-informed insight into whether robot automation is feasible for a given workcell. The future work could utilize optimal robot placement analysis~\cite{zielinski_robotgraffiti_2024} to support non-experts in complex robot placements and optimize the resulting trajectory from the path planner. Moreover, we have tested this tool for two automation tasks, but it should be extended to other applications with the proper constraints, e.g., welding by enforcing linear path planning between waypoints and a new interaction area tool allowing the user to pick a specific point rather than an area. Additionally, tested tasks had relatively simple robot placement objectives.}

On the negative side, we observed that users struggled with depth perception on a mobile device, indicating a need for simplifying some interface components, such as the bounding box setup. Some users also struggled to scan all important parts of the cell. For example, in task \textbf{(B)}, multiple participants failed to scan the top and back of the machine, resulting in failed feasibility analyses because the path planner found trajectories \say{through} the top of the machine due to the lack of a virtual representation of that part. Conversely, some users collected too many points within the machine and did not clean up the inside, leading to cases where the inside of the machine became unreachable due to floating points, resulting in no viable planning solution. {\color{black}Moreover, using non-iOS-powered mobile devices could lead to broader adoption of our approach but potentially introduce new issues.}

\section{Conclusions}
In this work, we present a novel reachability analysis tool designed for use by non-expert users. To demonstrate its effectiveness, we conducted expert user studies with five expert users (P1--P5), which helped us define two automation tasks with differing levels of difficulty (see \autoref{fig:formative_difficulty}). When using our prototype, the experts achieved at least a threefold reduction in the time required to complete feasibility studies compared to commonly used methods.

Next, through quantitative user studies with 22 non-experts, we have shown that the proposed method can be successfully used by non-experts, maintaining relatively similar cognitive load and usability regardless of the task complexity (see \autoref{fig:quantitative_nasa} and \autoref{tab:quantitative_fss_sus}). At the same time, users with little to no prior experience in automation were able to conduct feasibility studies efficiently, with time reductions of at least threefold compared to experts using traditional methods irrespective of task difficulty. This demonstrates the potential of the proposed method to bring significant savings for non-expert users who want to automate their existing production lines, particularly SMEs. By reducing the time and costs associated with the technical feasibility of automation, the proposed method promotes democratizing robotics automation. 

Despite these successes, we see room for improvement. Feedback and observations from participants indicated that certain AR interactions, such as placing a bounding box, require further refinement to enhance intuitiveness. This is reflected in the usability score, i.e., while an average score of 75.16 is considered very good, some interface refinements could lead to higher ratings. Additionally, while the used tool addresses the reachability aspect of feasibility analysis, other aspects, such as technical and economic feasibility, must also be considered for a comprehensive evaluation.

{\color{black}Moreover, while our solution is not designed to generate production-ready robot programs, we must focus on several future considerations. To further improve the quality of the reachability result and provide an estimate of a cycle time (a critical decision-making factor), the applied simulation for path planning needs to introduce dynamic (e.g., reduced reach due to the weight of the workpiece) and temporal calculations (e.g., time to open CNC machine). Furthermore, the solution's effectiveness depends on the quality of the user-provided input. Our toolbox enables users to improve the clarity of the captured complex scans even when LiDAR limitations are apparent. Users with knowledge about their application (but not about robot automation) can make educated decisions during the iterative scan refinement process to ensure the required elements are properly modeled.}

The mobile AR-based tool used in our work encourages users to interact with their environment in innovative and dynamic ways, highlighting the potential for mobile devices to support more complex and spatially aware tasks in robotics and other domains. By integrating physical movement and spatial awareness into the interaction model, the AR tool enhances user engagement with both the task and the environment. This approach not only improves the feasibility analysis process but also enriches the overall user experience, making it more immersive and intuitive. At the same time, we need to enhance this tool with more environmental data and clearer visual cues to ensure that both experts and non-experts can make well-informed decisions.

\section*{Ethical approval statement}
We followed the national regulations, which do not require ethics committee approval for non-medical user studies. We complied with IEEE’s PSPB Operations Manual by ensuring no harm to the participants, obtaining written and revocable consent, and anonymizing participants’ data.




\bibliographystyle{IEEEtran}
\bibliography{IEEEabrv,IEEEexample,mybibliography}



\end{document}